\documentclass[conference]{IEEEtran}
\IEEEoverridecommandlockouts
\usepackage{cite}
\usepackage{amsmath,amssymb,amsfonts}
\usepackage{algorithmic}
\usepackage{graphicx}
\usepackage{textcomp}
\usepackage{xcolor}
\usepackage{subfigure}
\usepackage{comment}

\def\BibTeX{{\rm B\kern-.05em{\sc i\kern-.025em b}\kern-.08em
    T\kern-.1667em\lower.7ex\hbox{E}\kern-.125emX}}
\begin{document}

\newcommand{\changed}[1]{\textcolor[rgb]{0.8,0.0,0.4}{{}{#1}}}
\newcommand{\todo}[1]{\textcolor[rgb]{0.0,0.8,0.4}{\small {~[TODO]~}{#1}}}

\title{Explainable Deep Reinforcement Learning\\Using Introspection in a Non-episodic Task}


\author{\IEEEauthorblockN{
Angel Ayala\,$^{1}$, 
Francisco Cruz\,$^{2,3}$, and
Bruno Fernandes\,$^{1}$, and
Richard Dezeley\,$^{2}$
}
\IEEEauthorblockA{\,
$^{1}$Escola Polit\'ecnica de Pernambuco, Universidade de Pernambuco, Recife, Brasil\\
$^{2}$School of Information Technology, Deakin University, Geelong, Australia\\
$^{3}$Escuela de Ingenier\'ia, Universidad Central de Chile, Santiago, Chile\\
Emails: \{aaam, bjtf\}@ecomp.poli.br, \{francisco.cruz, richard.dazeley\}@deakin.edu.au}
}

\maketitle

\begin{abstract}
Explainable reinforcement learning allows artificial agents to explain their behavior in a human-like manner aiming at non-expert end-users.
An efficient alternative of creating explanations is to use an introspection-based method that transforms Q-values into probabilities of success used as the base to explain the agent's decision-making process.
This approach has been effectively used in episodic and discrete scenarios, however, to compute the probability of success in non-episodic and more complex environments has not been addressed yet.
In this work, we adapt the introspection method to be used in a non-episodic task and try it in a continuous Atari game scenario solved with the Rainbow algorithm.
Our initial results show that the probability of success can be computed directly from the Q-values for all possible actions.
\end{abstract}


\section{Introduction}
Reinforcement learning (RL) \cite{sutton2018reinforcement} is a machine learning technique in which an autonomous agent acquires a new skill by interacting with the environment.
%
Lately, the development of deep learning methods, especially for image processing, has allowed the emergence of deep RL methods~\cite{churamani2020icub}.
Deep RL combines the use of deep neural networks and RL techniques~\cite{arulkumaran2017deep}.
In this approach, a deep learning architecture may process raw images as inputs giving the value function as output.
Therefore, Q-values can be obtained directly from unprocessed images that represent the state of the environment.

Empirical black-box models, as the case of deep RL, suffers the lack of interpretation in terms of their parameters~\cite{cruz2007indirect, cruz2010indirect}.
Therefore, it is not easy for a non-expert end-user to understand the behavior of a deep RL agent in simple domain-based language.
In this paper, we implement an introspection-based explainable method \cite{cruz2020explainable}.
The method is based on a mathematical transformation of the internal Q-values into probabilities of completing the intended task.
Therefore, the deep RL agent needs to look into the internal values, and no additional memory or processing is needed.
We have tested the introspection-based method in the Space Invaders environment, an Atari game scenario.


\section{Deep introspection-based approach}
Explanations from RL algorithms are useful to understand better why an action was taken in a particular situation by the agent.
Our researched approach estimates the probability of success $\hat{P}_s$ directly from the Q-values as the introspective knowledge of the agent's self-motivation.
This approach is efficient due to the direct numerical transformation from the Q-values into the probability of success and, therefore, no additional memory is needed to compute $\hat{P}_s$.
The Q-value in a given state-action pair for value-based reinforcement learning algorithms indicates how much future reward the agent may obtain.
Hence, using the Q-values, we can compute the probability of success $\hat{P}_s$ as follows:








\begin{equation}
    \label{eq:pSucess2}
    \hat{P}_s \approx \frac{1}{2} \cdot \log_{10} \frac{Q(s,a)}{R^S} + 1,
\end{equation}

where $Q(s,a)$ is the Q-value and $R^S$ the maximal reward obtained in any step.
As a probability value, we limited $\hat{P}_s \in [0, 1]$.
In order to compute the probability of success, in this work, we adapted Eq. \eqref{eq:pSucess2} from \cite{cruz2020explainable} to be used in non-episodic scenarios.
Therefore, we use the maximal possible reward obtained in any step $R^S$ instead of the maximal total reward $R^T$ as per episodic tasks.

The introspection approach is used along with the Rainbow deep RL algorithm \cite{hessel2018rainbow}, which integrates previous deep RL methods such as double Q-learning, prioritized replay, dueling networks, multi-step learning, distributional RL, and noisy nets including recommended parameters.






The network used as approximation function present a dueling network architecture from \cite{wang2016dueling} and adapted as \cite{hessel2018rainbow} with a shared representation $f_\xi(s)$, which is fed into a value stream $v_\eta$ with $N_\text{atoms}$ and into and advantage stream $a_\xi$, formulated as follows:

\begin{equation}
    \label{eq:RainbowAlgorithm}
    p_\theta^i(s,a) = \frac{\exp(v_\eta^i(\phi) + a_\psi^i(\phi, a) - \bar{a}_\psi^i(s))}
                            {\sum_j\exp(v_\eta^j(\phi) + a_\psi^j(\phi, a) - \bar{a}_\psi^j(s))},
\end{equation}

where $\phi = f_\xi(s)$ and $\bar{a}_\psi^i = \frac{1}{N_\text{actions}} \sum_{a'} a_\psi^i(\phi, a')$.
Additionally, it replaces all the linear layers with noisy layers equivalents as presented in \cite{fortunato2017noisy} and adapted in \cite{hessel2018rainbow}.

\section{Experimental Scenario}



Our approach has been tested in the Space Invaders Atari game environment.
The state space is composed of an RGB image with 210x160 pixels which is resized to 84x84 and converted to grayscale.
Additionally, 4 consecutive frames were stacked to emulate a spatio-temporal analysis according to the suggested by a DQN previous implementation \cite{mnih2015human}. 
The actions are in discrete domain, configured by 6 possible choices $\in[0, 5]$ with the following meaning,  
0: do nothing or no operation; 
1: fire; 
2: move to the right; 
3: move to the left;
4: action 2 plus 1; and
5: action 3 plus 1.

\begin{figure}
    \centering
    \subfigure[Initial state.]{\includegraphics[width=0.35\columnwidth]{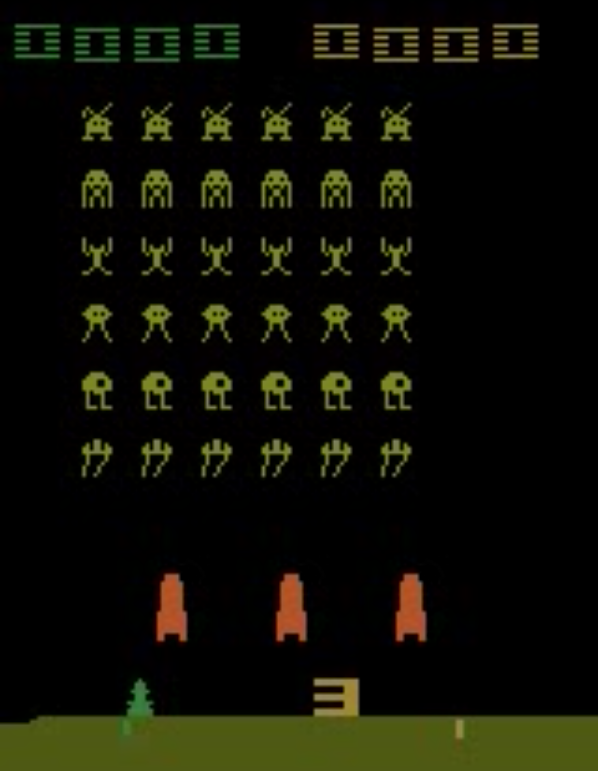}
        \label{fig:State0}
    }
    \quad
    \subfigure[Extra alien ship.]{\includegraphics[width=0.35\columnwidth]{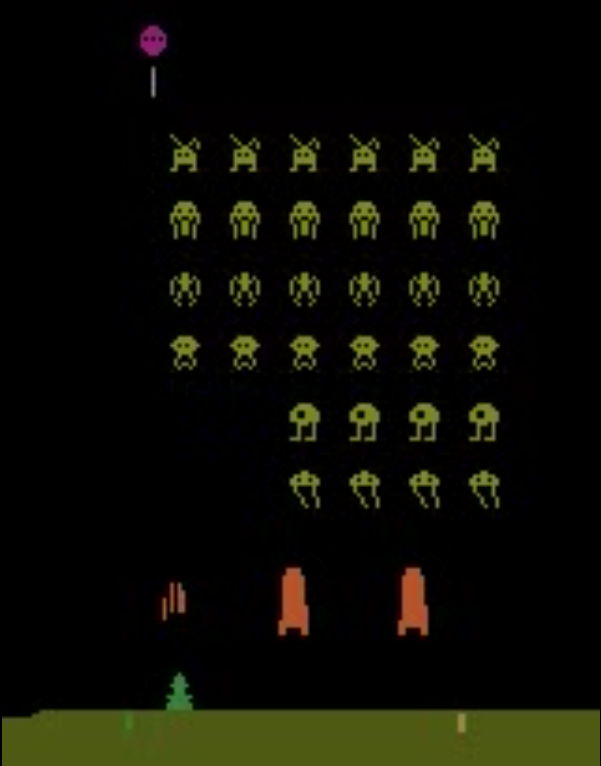}
        \label{fig:StateExtra}
    }
    \caption{Original observation for the Space Invaders arcade environment.
    }
    \label{fig:StateExamples}
\end{figure}

In this scenario, the agent must destroy a swarm of alien ships with three lives, i.e., it can receive a maximum of three alien bullets.
The agent starts at the bottom-left part of the frame as illustrated in Figure \ref{fig:State0} and must shoot to the alien swarm.
The reward obtained depends on the alien ship destroyed and its position in the swarm. It starts from 5 points for ships at the bottom row with an increment of 5 points for ships each row up, until 30 points for ships at the top row. 
Additionally, an extra alien ship appears randomly in the top part of the frame with a reward of 200 points, as shown in Figure \ref{fig:StateExtra}.

The agent was trained under 50 million steps in total using the Rainbow algorithm \cite{hessel2018rainbow} with all the recommended parameters. 
The agent was evaluated for each 1 million steps to check learning performance, and, therefore, the training process was split into 50 episodes.
Episode 0 was used to fill the prioritized replay memory with 80 thousand steps before starting the learning process.

\section{Discussion and Future Work}

\begin{figure}
    \centering
    \subfigure[Evaluation reward per episode.]{\includegraphics[width=0.5\linewidth]{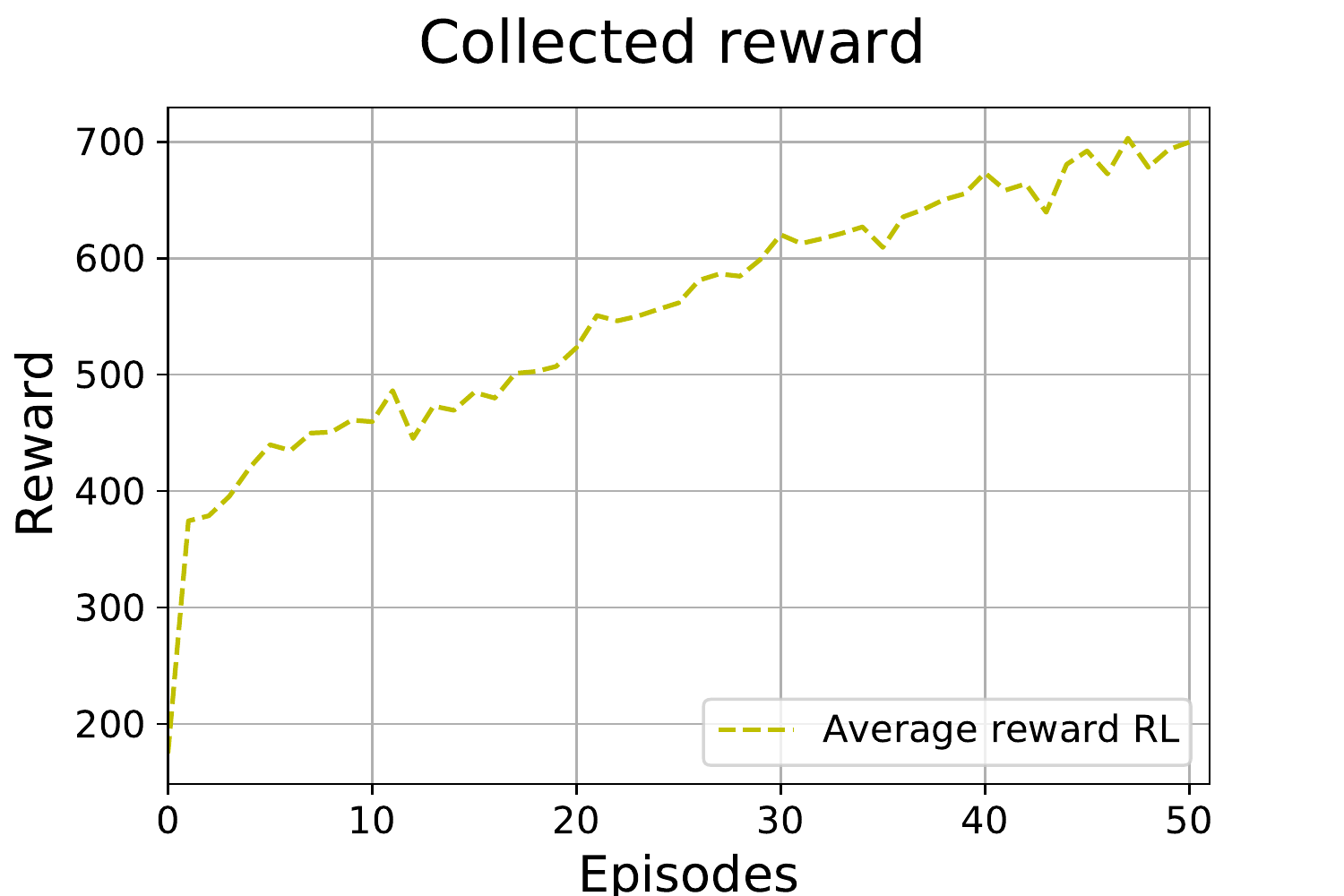}
        \label{fig:AvgRewards}
    }
    \subfigure[Probability of success.]{\includegraphics[width=0.5\linewidth]{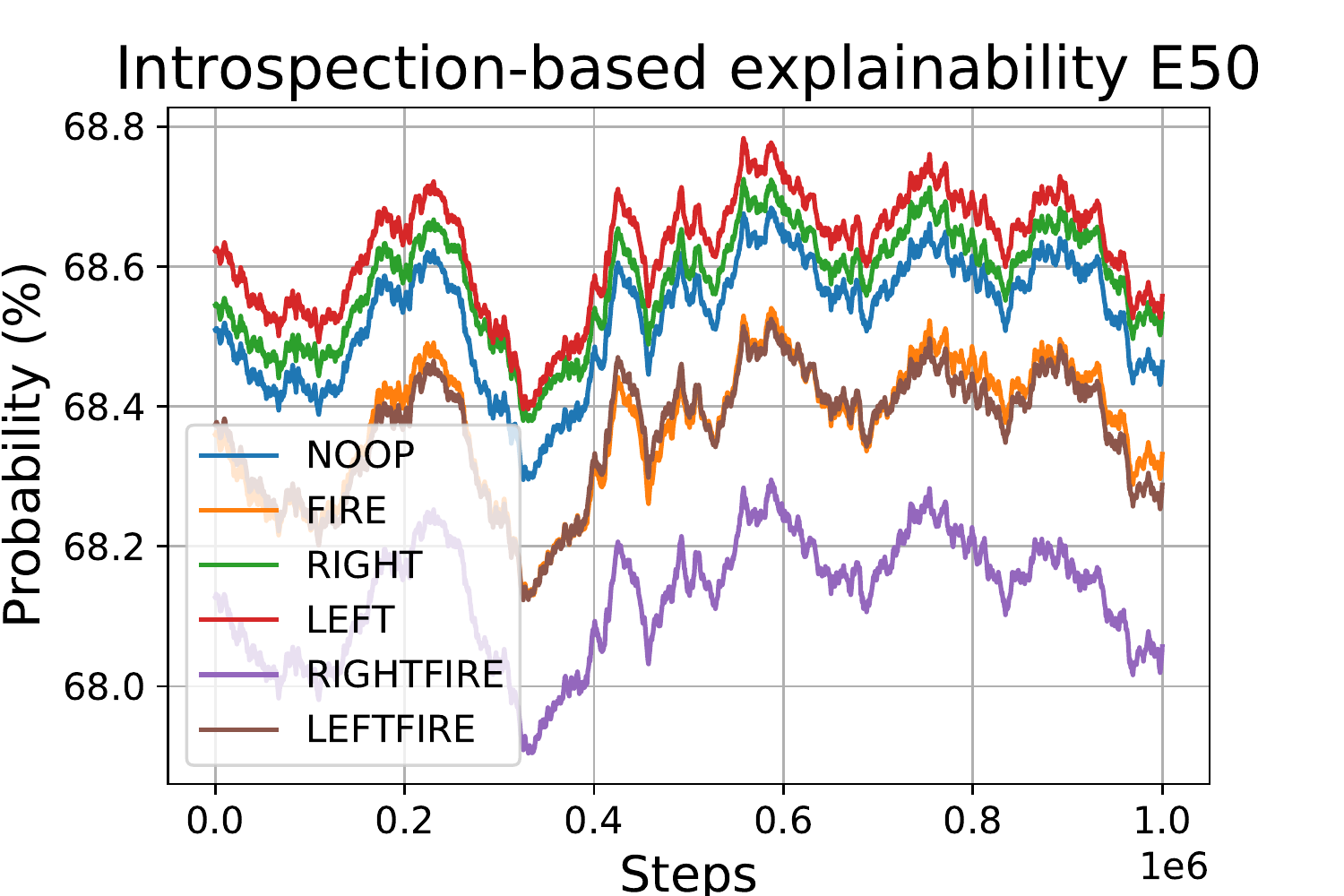}
        \label{fig:Instrospection}
    }
    \caption{Reward and probability of success from the initial state after training. 
    }
    \label{fig:Plots}
\end{figure}

After 50 epochs of training for 9 agents, all the agents were able to fulfill the task of destroying the swarm.  
In Figure \ref{fig:AvgRewards} can be observed that the agents during evaluation are able to get a high average reward, destroying at least one time the alien swarm, which summarizes 630 points. 
In the last episode (i.e., the last 1 million training steps), the average Q-values for all action varied between 6.85 and 7.15, which is meaningless to explain the agent behavior.
Therefore, using the introspection-based method as shown in Eq. \eqref{eq:pSucess2}, we have obtained the probabilities of success from the initial state.
Figure \ref{fig:Instrospection} shows the probability for each action during the last episode with values from 0.678 to 0.688.
As pointed out previously, these probabilities may be used as the base to explain the agent behavior 
more naturally to non-experts and provide a number easy to interpret than raw Q-values.

However, the computed probabilities are not so different among them as expected, and therefore, it may be difficult to use them to generate counterfactual explanations.
We hypothesize that this minor difference may be came from that in this complex scenario, a sequence of actions to complete the task may be elicited starting from any action.

Although the probability of success might provide a more understandable manner to explain the agent behavior \cite{cruz2020explainable}, it is still not clear to what extent.
Therefore, future work considers a user study to understand better if the probability of success is a valid metric to use as a base for explainable reinforcement learning.
This is especially relevant in complex scenarios in which the impact of an action, in terms of the intended goal, may not be clear easily by non-expert end-users.





\section*{Acknowledgment}
This work has been financed in part by the Coordenação de Aperfeiçoamento de Pessoal de Nível Superior - Brasil (CAPES)
- Finance Code 001, Fundação de Amparo a Ciência e Tecnologia do Estado de Pernambuco (FACEPE), Conselho Nacional de Desenvolvimento Científico e Tecnológico (CNPq) - Brazilian research agencies, and Universidad Central de Chile under the research project CIP2020013.


\begin{footnotesize}


\bibliographystyle{ieeetr}
\bibliography{biblio}

\end{footnotesize}

\end{document}